\documentclass[11pt,a4paper]{article}
\usepackage[hyperref]{acl2019}
\usepackage{times}
\usepackage{latexsym}
\usepackage{subcaption}
\usepackage{multirow,bigdelim}
\usepackage{amsmath}
\usepackage{rotating}
\usepackage{booktabs,xcolor}
\usepackage{xspace}
\usepackage{booktabs}
\usepackage{multirow}
\usepackage{todonotes}
\usepackage{url}

\newcommand{\bleu}{\texttt{BLEU}\xspace}
\newcommand{\meteor}{\texttt{METEOR}\xspace}
\newcommand{\base}{\texttt{base}\xspace}
\newcommand{\basesum}{\texttt{base+sum}\xspace}
\newcommand{\baseatt}{\texttt{base+att}\xspace}
\newcommand{\baseattobj}{\texttt{base+obj}\xspace}
\newcommand{\delib}{\texttt{del}\xspace}
\newcommand{\delibsum}{\texttt{del+sum}\xspace}
\newcommand{\delibatt}{\texttt{del+att}\xspace}
\newcommand{\delibattobj}{\texttt{del+obj}\xspace}
\newcommand{\rnd}{\texttt{RND}\xspace}
\newcommand{\amb}{\texttt{AMB}\xspace}
\newcommand{\pers}{\texttt{PERS}\xspace}
\newcommand{\comment}[1]{}

\aclfinalcopy 

\title{Distilling Translations with Visual Awareness}

\author{Julia Ive$^{1}$, Pranava Madhyastha$^{2}$ \and Lucia Specia$^{2}$\\
  $^{1}$DCS, University of Sheffield, UK\\
  $^{2}$Department of Computing, Imperial College London, UK \\
  {\tt j.ive@sheffield.ac.uk}\\
{\tt \{pranava,l.specia\}@imperial.ac.uk} \\}

\date{}

\begin{document}
\maketitle
\begin{abstract}
Previous work on multimodal machine translation has shown that visual information is only needed in very specific cases, for example in the presence of ambiguous words where the textual context is not sufficient. As a consequence, models tend to learn to ignore this information. 
We propose a translate-and-refine approach to this problem where images are only used by a second stage decoder. This approach is trained jointly to generate a good first draft translation and to improve over this draft by (i) making better use of the target language textual context (both left and right-side contexts) and (ii) making use of visual context. This approach leads to the state of the art results. Additionally, we show that it has the ability to recover from erroneous or missing words in the source language.
\end{abstract}

\section{Introduction}\label{sec:intro}

Multimodal machine translation (MMT) is an area of research that addresses the task of translating texts using context from an additional modality, generally static images. The assumption is that the visual context can help ground the meaning of the text and, as a consequence, generate more adequate translations. 
Current work has focused on datasets of images paired with their descriptions, which are crowdsourced in English and then translated into different languages, namely the Multi30K dataset \cite{elliott-etall_VL:2016}.

\begin{figure*}[t]
\small{
  \begin{subfigure}[c]{\textwidth}
  \vspace{1em}
    \begin{tabular}{c p{0.3cm}p{11cm}}
      \multirow{3}[15]{*}{\includegraphics[width=0.22\textwidth]{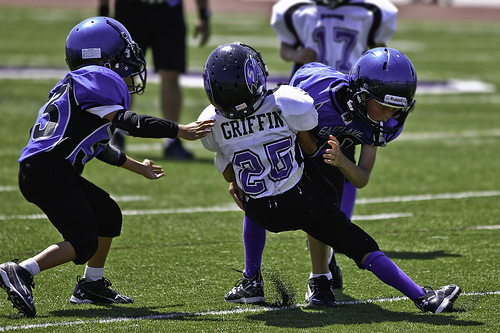}} & EN: & Three children in \underline{football} uniforms are playing \underline{football}. \\[1ex]
      & DE: & Drei Kinder in \underline{Fu{\ss}ball}trikots spielen \underline{Fu{\ss}ball}. \\[1ex] 
      & PE: & Drei Kinder in \underline{Footballtrikots}  spielen \underline{Football}. \\[1ex] 
  \end{tabular}
  \vspace{3em}
  \caption{\textbf{Ambiguous} word {\em football} translated as {\em soccer} ({\em Fu{\ss}ball})}
  \end{subfigure}
  \begin{subfigure}[c]{\textwidth}
  \vspace{1em}
    \begin{tabular}{c p{0.3cm}p{11cm}}
      \multirow{3}[15]{*}{\includegraphics[width=0.22\textwidth]{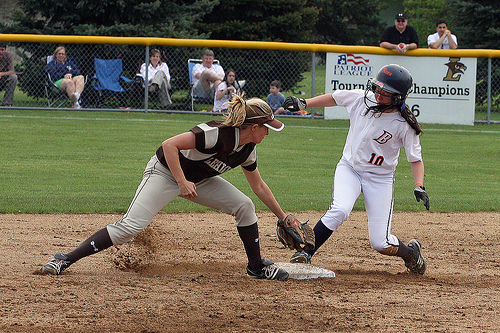}} & EN: & A baseball \underline{player} in a black shirt just tagged a player in a white shirt.\\[1ex]
      & DE: & \underline{Ein Baseballspieler} in einem schwarzen Shirt f\"{a}ngt \underline{einen Spieler} in einem wei{\ss}en Shirt.\\[1ex]
      & PE: & \underline{Eine Baseballspielerin} in einem schwarzen Shirt f\"{a}ngt \underline{eine Spielerin} in einem wei{\ss}en Shirt. \\[1ex]
  \end{tabular}
  \vspace{3em}
\caption{\textbf{Gender-neutral} word {\em player} translated as male player ({\em Spieler})}
  \end{subfigure}
  \begin{subfigure}[c]{\textwidth}
  \vspace{1em}
    \begin{tabular}{c p{0.3cm}p{11cm}}
      \multirow{3}{*}{\includegraphics[width=0.22\textwidth]{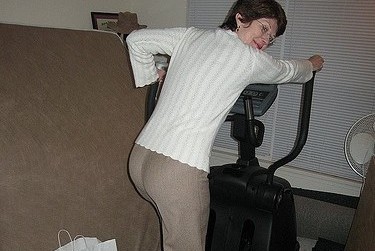}} & EN: & A woman wearing a white \underline{shirt} works out on an elliptical machine.\\[1ex]
      & DE: & Eine Frau in einem wei{\ss}en \underline{Shirt} trainiert auf einem Crosstrainer.\\[1ex]
      & PE: & Eine Frau in einem wei{\ss}en \underline{Pullover} trainiert auf einem Crosstrainer.\\[1ex]
  \end{tabular}
  \vspace{3em}
  \caption{\textbf{Inaccurate English} word {\em shirt} instead of {\em sweater} or {\em pullover}}
  \end{subfigure}}
 \caption{Examples of lexical and gender ambiguity, and inaccurate English description where post-edits (PE) required the image to correct human translation from English (EN) to German (DE).}\label{fig:examples}
\end{figure*}

Results from the most recent evaluation campaigns in the area \cite{elliott-EtAl:2017:WMT,BarraultEtAl:2018} have shown that visual information can be helpful, as humans generally prefer translations generated by multimodal models than by their text-only counterparts. However, previous work has also shown that images are only needed in very specific cases \cite{lala-EtAl:2018:WMT}. This is also the case for humans. \newcite{frank_elliott_specia_NLE:2018} (see Figure \ref{fig:examples}) concluded that visual information is needed by humans in the presence of the following: {\bf incorrect or ambiguous} source words and {\bf gender-neutral words} that need to be marked for gender in the target language. In an experiment where human translators were asked to first translate descriptions based on their textual context only and then revise their translation based on a corresponding image, they report that these three cases accounted for 62-77\% of the revisions in the translations in two subsets of Multi30K.  

Ambiguities are very frequent in Multi30K, as in most language corpora. \newcite{BarraultEtAl:2018} shows that in its latest test set, 358 (German) and 438 (French) instances (out of 1,000) contain at least one word that has more than one translation in the training set. However, these do not always represent a challenge for translation models: often the text context can easily disambiguate words (see baseline translation in Figure~\ref{table:gapped_ex}(a)); additionally, the models are naturally biased to generate the most frequent translation of the word, which by definition  is the correct one in most cases. 

The need to gender-mark words in a target language when translating from English can be thought of as a disambiguation problem, except that the text context is often less telling and the frequency bias plays ends up playing a bigger role (see baseline translation  in Figure~\ref{table:gapped_ex}(c)). This has been shown to be a common problem in neural machine translation \cite{D18-1334,DBLP:journals/corr/abs-1901-03116}, as well as in areas such as image captioning \cite{snowboard:2018} and co-reference resolution \cite{N18-2003}.  

Incorrect source words are common in Multi30K, as in many other crowdsourced or user-generated dataset. In this case the context may not be enough (see DE translation  in Figure~\ref{fig:examples}(c)). We posit that models should be robust to such a type of noise and note that similar treatment would be required for out of vocabulary (OOV) words, i.e. correct words that are unknown to the model. 

We propose an approach that takes into account the strengths of a text-only baseline model and only {\bf refines} its translations when needed. Our approach is based on {\bf deliberation networks} \cite{xia2017deliberation} to jointly learn to generate draft translations and refine them based on left and right side target context as well as structured visual information. This approach outperforms previous work. 

In order to further probe how well our models can address the three problems mentioned above,
we perform a controlled experiment where we minimise the interference of the frequency bias by masking ambiguous and gender-related words, as well as randomly selected words (to simulate noise and OOV). This experiment shows that our multimodal refinement approach outperforms the text-only one in more complex linguistic setups. 

Our main {\bf contributions} are: (i) a novel approach to MMT based on deliberation networks and structured visual information which gives state of the art results (Sections \ref{ssec:img-models} and \ref{ssec:standard_setup}); (ii) a frequency bias-free investigation on the need for visual context in MMT (Sections \ref{sec:source_degradation} and \ref{ssec:gaps_results}); and (iii) a thorough investigation on different visual representations for transformer-based architectures (Section \ref{ssec:img-info}).

\section{Related work}\label{sec:related}

\paragraph{MMT:} Approaches to MMT vary with regards to how they represent images and how they incorporate this information in the models. Initial approaches use RNN-based sequence to sequence models \cite{BahdanauEtAl:2015} enhanced with a single, global image vector, extracted as one of the layers of a CNN trained for object classification~\cite{he2016deep}, often the penultimate or final layer. 

The image representation is integrated into the MT models by initialising the encoder or  decoder~\cite{ElliottEtAl:2015,CaglayanEtAl:2017,MadhyasthaEtAl:2017}; element-wise multiplication with the source word annotations~\cite{CaglayanEtAl:2017}; or  projecting the image representation and encoder context  to a common space to initialise the  decoder~\cite{CalixtoLiu:2017}. \newcite{ElliottKadar:2017} and \newcite{HelclEtAl:2018} instead model the source sentence and reconstruct the image representation jointly via multi-task learning. 

An alternative way of exploring image representations is to have an attention mechanism ~\cite{BahdanauEtAl:2015} on the output of the last convolutional layer of a CNN~\cite{XuEtAl:2015}. 
The layer represents the activation of $K$ different convolutional filters on evenly quantised $N \times N$ spatial regions of the image. \newcite{CaglayanEtAl:2017}  learn the attention weights for both source text and visual encoders, while  \newcite{CalixtoEtAl:2017,DelbrouckDupont:2017} combine both attentions independently via a gating scalar, and   \newcite{LibovickyHelcl:2017,HelclEtAl:2018} apply a hierarchical attention distribution over two projected vectors where the attention for each is learnt independently.

\newcite{HelclEtAl:2018} is the closest to our work: we also use a doubly-attentive transformer architecture and explore spatial visual information. However, we differ in two main aspects (Section \ref{sec:model}): (i) our approach explores additional textual context through a second pass decoding process and uses visual information only at this stage, and (ii) in addition to convolutional filters we use  object-level visual information. The latter has only been explored to generate a single global representation \cite{GronroosEtAl:2018} and used for example to initialise the encoder \cite{HuangEtAl:2016}.
We note that translation refinement is different translation re-ranking from a text-only model based on image representation~\cite{ShahEtAl:2016,HitschlerEtAl:2016,lala-EtAl:2018:WMT}, since the latter assumes that the correct translation can already be produced by a text-only model.  

~\citet{caglayan2019probing} investigate the importance and the contribution of multimodality for MMT. They perform careful experiments by using input degradation and observe that, specially under limited textual context, multimodal models exploit the visual input to generate better translations.~\citet{caglayan2019probing} also show that MMT systems exploit visual cues and obtain correct translations even with typographical errors in the source sentences. In this paper, we build upon this idea and investigate the potential of visual cues for refining translation.

\paragraph{Translation refinement:} The idea of treating machine translation as a two step approach dates back to statistical models, e.g. in order to improve a draft sentence-level translation by exploring document-wide context through hill-climbing for local refinements \cite{D12-1108}. Iterative refinement approaches have also been proposed that start with a draft translation and then predict discrete substitutions based on an attention mechanism 
 \cite{DBLP:journals/corr/NovakAG16}, or using non-autoregressive methods with a focus on speeding up decoding \cite{D18-1149}. Translation refinement can also be done through learning a separate model for automatic post-editing \cite{C16-1172,I17-1013,chatterjee-EtAl:2018:WMT}, but this requires additional training data with draft translations and their correct version.

An interesting approach is that of deliberation networks, which jointly train an encoder and first and second stage decoders \cite{xia2017deliberation}. The second stage decoder has access to both left and right side context and this has been shown to improve translation \cite{xia2017deliberation,hassan2018achieving}. We follow this approach as it offers a very flexible framework to incorporate additional information in the second stage decoder.

\section{Model}
\label{sec:model}

We base our model on the {\bf transformer architecture}~\cite{vaswani2017attention} for neural machine translation. Our implementation is a multi-layer encoder-decoder architecture that uses the \texttt{tensor2tensor}\footnote{\url{https://github.com/tensorflow/tensor2tensor}}~\cite{vaswani2018tensor2tensor} library. The encoder and decoder blocks are as follows: 

\paragraph{Encoder Block} ($\mathcal{E}$):  The encoder block comprises of $6$ layers, with each containing two sublayers of multi-head self-attention mechanism followed by a fully connected feed forward neural network. We follow the standard implementation and employ residual connections between each layer, as well as layer normalisation. The output of the encoder forms the encoder memory which consists of contextualised representations for each of the source tokens ($M_{\mathcal{E}}$).

\paragraph{Decoder Block} ($\mathcal{D}$): The decoder block also comprises of $6$ layers. It contains an additional sublayer which performs multi-head attention over the outputs of the encoder block. Specifically, decoding layer $d_{l_i}$ is the result of a) multi-head attention over the outputs of the encoder which in turn is a function of the encoder memory and the outputs from the previous layer: $A_{\mathcal{D}{\rightarrow}\mathcal{E}} = f(M_{\mathcal{E}}, d_{l_{i-1}})$ where, the keys and values are the encoder outputs and the queries correspond to the decoder input, and b) the multi-head self attention which is a function of the generated outputs from the previous layer: $A_{\mathcal{D}} = f(d_{l_{i-1}})$.

\subsection{Deliberation networks} 
\begin{figure}[!htbp] 
\center{\includegraphics[scale=0.09]{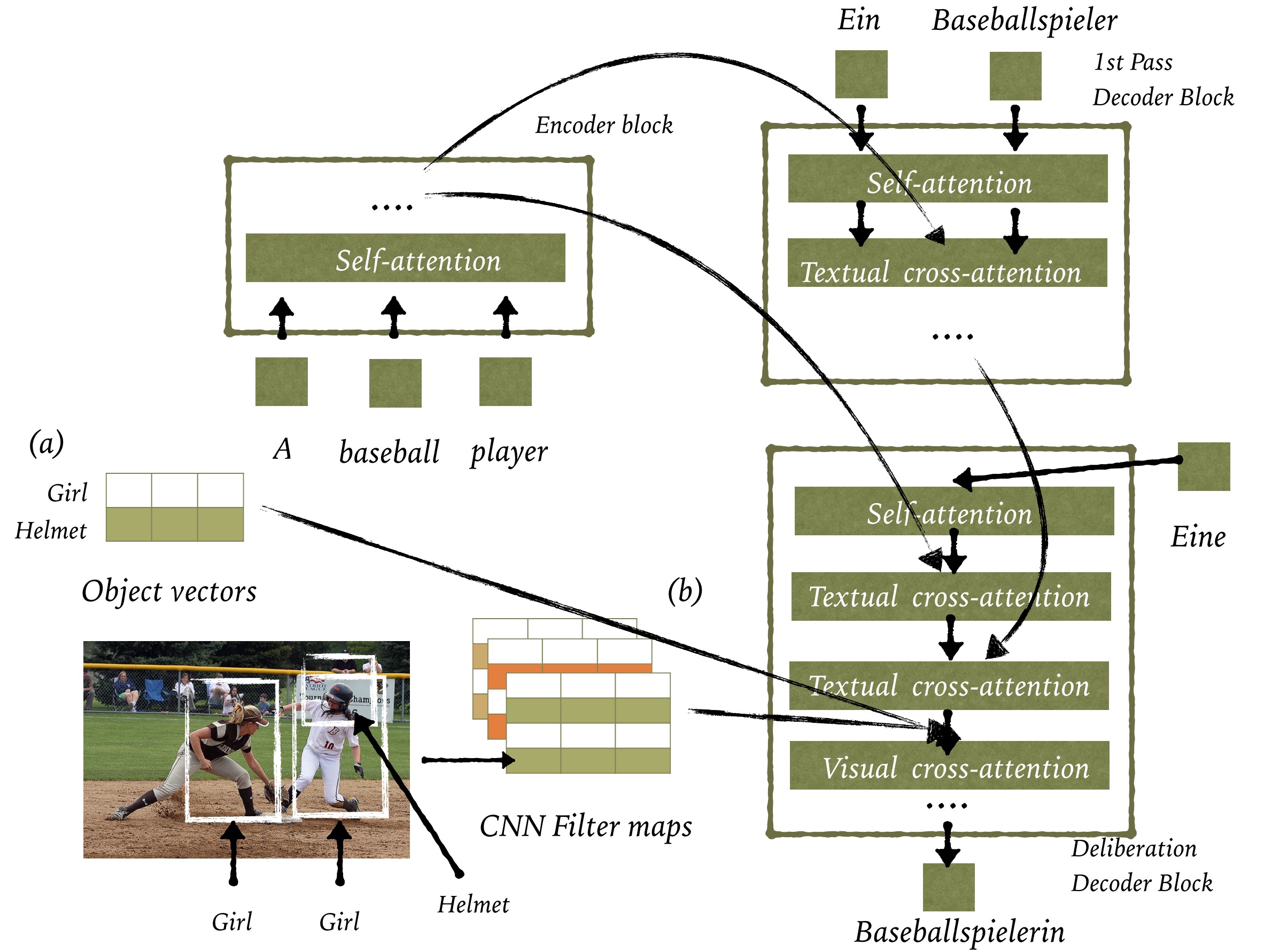}}
\caption{Our deliberation architecture: The second-pass
 decoder is conditioned on the source and samples output from the first-pass decoder. The second-pass decoder has access to (a) the object based features represented by embeddings, or (b) spacial image features.}
\label{figure:delib}
\end{figure}

Deliberation
networks~\cite{hassan2018achieving,xia2017deliberation} build on the standard
sequence to sequence architecture to add an additional decoder block (in our case, with $3$ layers  -- see Figure~\ref{figure:delib}). The additional
decoder (also referred to as second-pass decoder) is conditioned on the source and sampled outputs from the standard transformer decoder (the first-pass decoder). More concretely, the second-pass decoder ($\mathcal{D}^{'}$) at layer $d^{'}_{l}$ consists of $A_{\mathcal{D}^{'}}$ , $A_{\mathcal{D^{'}}{\rightarrow}\mathcal{E}}$, $A_{\mathcal{D^{'}}{\rightarrow}\mathcal{D}}$, where, 
$A_{\mathcal{D}^{'}}$ and $A_{\mathcal{D^{'}}{\rightarrow}\mathcal{E}}$ is similar to the standard deliberation architecture multi-head attention over the encoder memory and self attention respectively while, $A_{\mathcal{D^{'}}{\rightarrow}\mathcal{D}}$ is the multi-head attention over outputs $O_{d}$ from the first-pass decoder ($\mathcal{D}$) ($A_{\mathcal{D}{\rightarrow}\mathcal{E}} = f(O_{d}, d^{'}_{l_{i-1}})$).\footnote{In the implementation we used, the deliberation network trains 345M parameters, as compared to the Transformer with 210M parameters.} In our experiments, we obtain samples as a set of translations from the first-pass decoder using beam-search. Given a translation candidate, $O_{d}$ consists of the first-pass decoder's hidden layer before softmax concatenated with the embeddings of the resultant words.

\subsection{Multimodal transformer \& deliberation}\label{ssec:img-models}

Our multimodal transformer models follow one of the two formulations below for conditioning translations on image information:

\paragraph{Additive image conditioning (AIC):} A
projected image vector is added to each of the outputs of the encoder. The projections matrices are parameters that are jointly learned with the model.

\paragraph{Attention over image features (AIF):} The model attends over image features, as in \newcite{HelclEtAl:2018}, where the decoder block now contains an additional cross-attention sub-layer 
$A_{\mathcal{D^{'}}{\rightarrow}\mathcal{V}}$ which attends to the visual information ($\mathcal{V}$). The keys and values correspond to the visual information.

Within the deliberation network framework, based on the previously discussed observation (Section \ref{sec:intro}) that images are only needed in a small number of cases, we propose to add visual cross-attention only to the second-pass decoder block (see Figure~\ref{figure:delib}).

\subsection{Image features}\label{ssec:img-info}
Motivated by previous work that indicates the importance of structured information from images~\cite{CaglayanEtAl:2017,wang2018object,madhyastha2018end}, we focus on structural forms of image representations, including the spatially aware feature maps from CNNs and information extracted from automatic  object detectors. 

\textbf{Spatial image features:} We use spatial feature maps from the last convolutional layer of a pre-trained ResNet-50~\cite{he2016deep} CNN-based image classifier for every image.\footnote{Provided at  \url{http://statmt.org/wmt18/multimodal-task.html.}}  These feature maps contain output activations for various filters while preserving spatial information. They have been used in various vision to language tasks including image captioning~\cite{XuEtAl:2015} and  multimodal machine translation (Section \ref{sec:related}). 
Our formulation for the integration of these features into the deliberation network is shown in Figure~\ref{figure:delib}, setup (b). We use the the AIF setup and refer to models that use the representation as \texttt{att}. 

\textbf{Object-based image features:}
We use a bag-of-objects representation where the objects are obtained using an off-shelf object detector~\cite{kuznetsova2018open} based on the Open Images dataset. This representations is a sparse $545$-dimensional vector with the frequency of each (545) given object in an image. This is inspired by previous research that investigates the potential of object-based information for vision to language tasks~\cite{mitchell2012midge,wang2018object}. 
We use the the AIC setup and refer to models that use the representation as  \texttt{sum}. 

\textbf{Object-based embedding features:}
The bag-of-objects representations makes it hard to  exploit object-to-object similarity, since visual representations of different objects can be very different. To mitigate this, we propose a simple extension using bag-of-object {\em embeddings}. We represent each object using the pre-trained  GLoVe-based~\cite{pennington2014glove} $50$-dimensional word vectors for their categories (e.g. {\em woman}). We use the the AIF based setup and  refer to models that use the representation as ~\texttt{obj} (Figure~\ref{figure:delib} setup (a)).
 
\section{Experimental settings}\label{sec:exps}

\subsection{Data}\label{ssec:settings}
We build and test our MMT models on the  \textbf{Multi30K} dataset~\cite{elliott-etall_VL:2016}. 
Each image in Multi30K contains one English (EN) description taken from Flickr30K~\cite{YoungEtAl:2014} and human translations into German (DE), French (FR) and Czech~\cite{SpeciaEtAl:2016,elliott-EtAl:2017:WMT,BarraultEtAl:2018}. The dataset contains 29,000 instances for training, 1,014 for development, and 1,000 for test. We only experiment with German and French, which are languages for which we have in-house expertise for the type of analysis we present. In addition to the official Multi30K test set (test 2016), we also use the test set from the latest WMT evaluation competition, test 2018 \cite{BarraultEtAl:2018}.\footnote{The pre-processed datasets  provided by the organisers were used without additional pre-processing.}

\subsection{Degradation of source}\label{sec:source_degradation}

In addition to using the Multi30K dataset as is (standard setup), we probe the ability of our models to address the three linguistic phenomena where additional context has been proved important (Section \ref{sec:intro}): ambiguities, gender-neutral words and noisy input. In a controlled experiment where we aim to remove the influence of frequency biases, we degrade the source sentences by masking words through three strategies to replace words by a placeholder: random source words, ambiguous source words and gender unmarked source words. The procedure is applied to the train, validation and test sets. For the resulting dataset generated for each setting, we compare  models having access to text-only context versus additional text and multimodal contexts. We seek to get insights into the contribution of each type of context to address each type of degradation.

\paragraph{Random content words}
In this setting (\rnd) we simulate erroneous source words by randomly dropping source content words. We first tag the entire source sentences using the \texttt{spacy} toolkit~\cite{honnibal2017spacy} and then drop nouns, verbs, adjectives and adverbs and replace these with a default \texttt{BLANK} token. By focusing on content words, we differ from previous work that suggests that neural machine translation is robust to non-content word noise in the source~\cite{klubivcka2017fine}. 

\paragraph{Ambiguous words}
In this setting (\amb), we rely on the MLT dataset~\cite{lala-EtAl:2018:WMT} which provides a list of source words with multiple translations in the Multi30k training set. We replace ambiguous words with the  \texttt{BLANK} token in the source language, which results in two language-specific datasets.

\paragraph{Person words} In this setting (\pers), we use the Flickr Entities dataset~\cite{plummer2015flickr30k} to identify all the words that were annotated by humans as corresponding to the category person.\footnote{We pre-processed the initial dataset to remove noise. We also add the gender-marked pronouns \emph{he}, \emph{she}, \emph{her} and \emph{his} to the person word list.} We then replace such source words with the  \texttt{BLANK} token.

The statistics of the resulting datasets for the three degradation strategies are shown in Table~\ref{table:gapped_stat}. We note that \rnd and \pers are the same for language pairs as the degradation only depends on the source side, while for \amb the words replaced depend on the target language.

\begin{table}[!h]
\begin{center}
\scalebox{0.73}{
\begin{tabular}{ccc}
\toprule
setup & \% sent. & avg. blanks per sent. \\ \midrule
\rnd & 100 & 1.5\\
\amb DE & 83 & 2 \\
\amb FR & 77 & 1.8\\
\pers & 92 & 1.6 \\
\bottomrule
\end{tabular}}
\end{center}
\caption{\label{table:gapped_stat} Statistics of datasets after applying source degradation strategies}
\end{table}

\subsection{Models}
Based on the models described in  Section~\ref{sec:model} we experiment with eight variants: (a)~baseline transformer model (\base); (b)~\base with AIC (\basesum); (c)~\base with AIF using spacial (\baseatt) or object based (\baseattobj) image features; (d)~standard deliberation model (\delib); (e)~deliberation models enriched with image information: \delibsum, \delibatt and \delibattobj.

\subsection{Training} 
In all cases, we optimise our models with cross entropy loss. For deliberation network models, we first train the standard transformer model until convergence, and use it to initialise the encoder and first-pass decoder. For each of the training samples, we follow~\cite{xia2017deliberation} and obtain a set of $10$-best samples from the first pass decoder, with a beam search of size $10$. We use these as the first-pass decoder samples. We use Adam as optimiser~\cite{kingma2014adam} and train the model until convergence.\footnote{We built on the \texttt{tensor2tensor} implementation of deliberation nets in \url{https://github.com/ustctf/delibnet} using the \texttt{transformer\_big} parameters with a learning rate of 0.05 with 8K warmup steps for both the first and the second-pass decoders, and early stopping with the patience of 10 epochs based on the validation \bleu{} score.}  

\begin{table}[!h]
\begin{center}
\scalebox{0.73}{
\begin{tabular}{lc@{\hspace{1cm}}ll@{\hspace{1cm}}lc}
\toprule
& & \multicolumn{2}{l}{test 2016} & \multicolumn{2}{l}{test 2018} \\ 
\cline{3-6}
& model & M & B & M & B \\ \midrule
\multirow{8}{*}{\rotatebox{90}{DE}}
& MMT \tiny{\cite{HelclEtAl:2018}} & 53.1 & 38.4 & - & - \\
\cline{2-6}
& \base & 54.5 & 36.4 & 45.0 & 26.5 \\
& \basesum & 54.2 & 35.9 & 45.0 & 26.4 \\
& \baseatt & 54.5 & 36.9 & 45.3 & 27.2 \\
& \baseattobj & 54.5 & 36.4 & 45.0 & 26.7 \\
\cline{2-6}
& \delib & \bf 55.5* & 37.7 & \bf 46.3* & 27.7\\
& \delibsum & \bf 55.2* & 37.3 & \bf 46.3* & 27.7 \\
& \delibatt & \bf 55.1* & 37.2 & \bf 46.1* & 27.4 \\
& \delibattobj & \bf 55.6* & 38.0 & \bf 46.5* & 27.6\\
\midrule
\multirow{8}{*}{\rotatebox{90}{FR}}
& MMT \tiny{\cite{HelclEtAl:2018}} & 75.0 & 60.6 & - & - \\
\cline{2-6}
& \base & 73.7 & 59.0 & 56.4 & 37.0 \\
& \basesum & 73.9 & 59.2 & 56.6 & 37.1 \\
& \baseatt & 73.5 & 58.7 & 56.1 & 36.2 \\
& \baseattobj & 72.9 & 57.3 & 55.8 & 36.3 \\
\cline{2-6}
& \delib & \bf 74.6* & 60.1 & \bf 57.2* & 37.8 \\
& \delibsum & \bf 74.3* & 59.6 & \bf 56.9* & 37.2\\
& \delibatt & 73.7$\dagger$ & 59.2 & 56.3$\dagger$ & 36.9 \\
& \delibattobj & \bf 74.4* & 59.8 & \bf 57.0* & 37.4\\
\bottomrule
\end{tabular}}
\end{center}
\caption{\label{table:main_res} Results for the test sets 2016 and 2018. M denotes \meteor{}, B -- \bleu{}; * marks statistically significant changes for \meteor (p-value $\leq 0.05$) as compared to \base{}, $\dagger$ -- as compared to \delib{}. Bold highlights  statistically significant improvements. We report previous state of the art results for multimodal models from \cite{HelclEtAl:2018}.}
\end{table}
 
\begin{figure*}[!h]
  \begin{subfigure}[c]{\textwidth}
  \vspace{1em}
  \begin{small}
    \begin{tabular}{c p{1.5cm}p{10cm}}
      \multirow{3}[15]{*}{\includegraphics[width=0.22\textwidth]{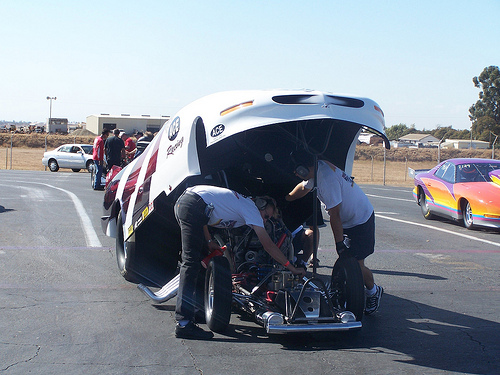}} & EN: & Two men work under the hood of a white \underline{race car}. \\[1ex]
      & \baseatt: & Zwei M{\"a}nner arbeiten unter der Motorhaube eines wei{\ss}en \underline{Rennens}.\\[1ex]
      & \delib: & Zwei M{\"a}nner arbeiten unter der Motorhaube eines wei{\ss}en \underline{Autos}.\\[1ex]
      & \delibattobj: & Zwei M{\"a}nner arbeiten unter der Motorhaube eines wei{\ss}en \underline{Rennwagen}.\\[1ex]
      & DE: & Zwei M{\"a}nner arbeiten unter der Haube eines wei{\ss}en \underline{Rennautos}. \\[1ex]
  \end{tabular}
  \end{small}
  \caption{ \baseatt translates \textit{race car} with \textit{Rennen} (race), \delib with \textit{Auto} (car) and \delibattobj with \textit{Rennwagen} (race car). \\ \textbf{Objects}: land, vehicle, car, wheel}
  \end{subfigure}
  \begin{subfigure}[c]{\textwidth}
  \vspace{1em}
    \begin{small}
    \begin{tabular}{c p{1.7cm}p{9.6cm}}
      \multirow{3}[15]{*}{\includegraphics[width=0.22\textwidth]{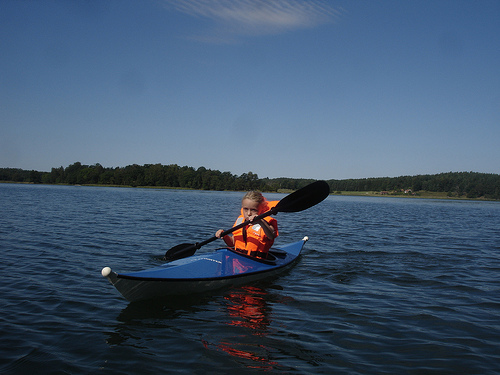}} & EN: & A young child holding an oar \underline{paddling} a blue kayak \underline{in a body of water}. \\[1ex]
      & \baseatt: & Un jeune enfant tenant une rame dans un kayak bleu. \\[1ex]
      & \delib: &  Un jeune enfant tenant une rame dans un kayak bleu \underline{sur un plan d'eau}. \\[1ex]
      & \delibattobj: & Un jeune enfant tenant une rame dans un kayak bleu \underline{pagayant} \underline{sur un plan d'eau}. \\[1ex]
      & FR: & Un jeune enfant avec une rame \underline{pagayant} dans un kayak bleu \underline{sur un plan d'eau}. \\[1ex]
  \end{tabular}
  \end{small}
  \caption{\delib and \delibattobj translate \textit{in a body of water} with \textit{sur un plan d'eau} (on a body of water), missing in \baseatt. \delibattobj translates the word \textit{paddling} with \textit{pagayant} (paddling). \textbf{Objects}: paddle, canoe}
  \end{subfigure}
\caption{\label{fig:main_ex} Examples of improvements of \delib and \delibattobj over \baseatt for test set 2016 for French and German. Underlined words represent some of the improvements.}
\end{figure*}

\section{Results}\label{ssec:results}

In this section we present results of our experiments, first in the original dataset without any source degradation (Section \ref{ssec:standard_setup}) and then in the setup with various source degradation strategies (Section \ref{ssec:gaps_results}).

\subsection{Standard setup}\label{ssec:standard_setup}

Table~\ref{table:main_res} shows the results of our main experiments on the 2016 and 2018 test sets for French and German. We use Meteor \cite{denkowski:lavie:meteor-wmt:2014} as the main metric, as in the WMT tasks \cite{BarraultEtAl:2018}. We compare our transformer baseline to transformer models enriched with image information, as well as to the deliberation models, with or without image information. 

We first note that our multimodal models achieve the state of the art performance for transformer networks (constrained models) on the English-German dataset, as compared to \cite{HelclEtAl:2018}. 
Second, our deliberation models lead to significant improvements over this baseline across test sets (average $\Delta\meteor=1$, $\Delta\bleu=1$). 

Transformer-based models enriched with image information (\basesum, \baseatt and \baseattobj), on the other hand,  show no major improvements with respect to the \base performance. This is also the case for deliberation models with image information (\delibsum, \delibatt, \delibattobj), which do not show significant improvement over the vanilla deliberation performance (\delib).

However, as it has been shown in the WMT shared tasks on MMT \cite{SpeciaEtAl:2016,elliott-EtAl:2017:WMT, BarraultEtAl:2018}, automatic metrics often fail to capture nuances in translation quality, such as, the ones we expect the visual modality to help with, which -- according to human perception -- lead to better translations.  To test this assumption in our settings, we performed human evaluation involving professional translators and native speakers of both French and German (three annotators). 

The annotators were asked to rank randomly selected test samples according to how well they convey the meaning of the source, given the image (50 samples per language pair per annotator). For each source segment, the annotator was shown the outputs of three systems: \baseatt, the current MMT state-of-the-art \cite{HelclEtAl:2018}, \delib and \delibattobj. A rank could be assigned from 1 to 3, allowing  ties~\cite{W17-4717}. Annotators could assign zero rank to all translations if they were judged incomprehensible.
Following the common practice in WMT \cite{W17-4717}, each system was then assigned a score which reflects the proportion of times it was judged to be better or equal other  systems.

Table~\ref{table:human_res} shows the human evaluation results. They are consistent with the automatic evaluation results when it comes to the preference of humans towards the deliberation-based setups, but show a more positive outlook regarding the addition of visual information (\delibattobj over \delib) for French.

\begin{table}[!h]
\begin{center}
\scalebox{0.73}{
\begin{tabular}{c c c c}
\toprule
lang & \baseatt & \delib & \delibattobj \\ \midrule
DE & 0.35	& {\bf 0.62} & 0.59  \\
FR & 0.41 & 0.6 & {\bf 0.67} \\
\bottomrule
\end{tabular}}
\end{center}
\caption{\label{table:human_res} Human ranking results: normalised rank (micro-averaged). Bold highlights best results.}
\end{table}

Manual inspection of translations suggests that deliberation setups tend to improve both the grammaticality and adequacy of the first pass outputs. For German, the most common modifications performed by the second-pass decoder are substitutions of adjectives and verbs (for test 2016, 15\% and 12\% respectively, of all the edit distance operations). Changes to adjectives are mainly grammatical, changes to verbs are contextual (e.g., changing \textit{laufen} to \textit{rennen}, both verbs mean run, but the second refers to running very fast). For French, 15\% of all the changes are substitutions of nouns (for test 2016). These are again very contextual. For example, the French word \textit{travailleur} (worker) is replaced by \textit{ouvrier} (manual worker) in the contexts where tools, machinery or buildings are mentioned. For our analysis we used again \texttt{spacy}.

The information on detected objects is particularly helpful for specific adequacy issues. Figure~\ref{fig:main_ex} demonstrates some such cases. In the first case, the \baseatt model misses the translation of \textit{race car}: the German word \textit{Rennen} translates only the word \textit{race}. \delib introduces the word \textit{car} (\textit{Auto}) into the translation. Finally, \delibattobj correctly translates the expression \textit{race car} (\textit{Rennwagen}) by exploiting the object information. For French, \delib translates the source part \textit{in a body of water}, missing from the \baseatt translation. \delibattobj additionally translated the word \textit{paddling} according to the detected object Paddle.

\begin{table*}[!ht]
\begin{center}
\scalebox{0.73}{
\begin{tabular}{lc@{\hspace{1cm}}ll@{\hspace{0.5cm}}ll@{\hspace{1.2cm}}ll@{\hspace{0.5cm}}ll@{\hspace{1.2cm}}ll@{\hspace{0.5cm}}ll}
\toprule
& & \multicolumn{3}{c}{\rnd} & & \multicolumn{3}{c}{\amb} & & \multicolumn{3}{c}{\pers} \\ 
\cline{3-14}
& & \multicolumn{2}{l}{test 2016} & \multicolumn{2}{l}{test 2018} & \multicolumn{2}{l}{test 2016} & \multicolumn{2}{l}{test 2018} & \multicolumn{2}{l}{test 2016} & \multicolumn{2}{l}{test 2018} \\
\cline{3-14}
& model & M & B & M & B & M & B & M & B & M & B & M & B \\ \midrule
\multirow{4}{*}{\rotatebox{90}{DE}}
& \base & 45.6 & 27.1 & 37.7 & 20.0 & 48.4 & 30.1 & 38.9 & 21.0 & 47.0 & 28.6 & 40.3 & 22.2\\
& \delib & 44.6* & 25.1 & 36.8* & 18.1 & 47.7 & 29.0 & 38.0* & 19.0 & 47.5 & 29.0 & 40.9 & 22.0 \\
& \delibsum & 45.7$\dagger$ & 27.2 & 38.1$\dagger$ & 19.9 & 46.9*$\dagger$ & 27.9 &  37.2*$\dagger$ & 18.7 & \bf 48.1* & 29.8 & \bf 41.1* & 22.4\\
& \delibattobj & \bf 46.5*$\dagger$ & 28.1 & \bf 39.0*$\dagger$ & 20.7 & \bf 49.8*$\dagger$ & 31.3 & \bf 40.0*$\dagger$ & 21.3 & \bf 48.1* & 29.4 & \bf 41.6*$\dagger$ & 23.4 \\\midrule
\multirow{4}{*}{\rotatebox{90}{FR}}
& \base & 59.3 & 43.4 & 46.3 & 28.1 & 66.4 & 51.2 & 49.2 & 30.4 & 63.9 & 48.6 & 50.3 & 31.7\\
& \delib & \bf 61.0* & 45.3 & \bf 47.1* & 28.4 & \bf 67.3* & 52.2 & \bf 50.2* & 31.3 & \bf 64.5* & 49.3 & \bf 51.2* & 32.4\\
& \delibsum & \bf 60.4* & 44.4 & \bf 47.5* & 29.3 & \bf 67.7* & 52.8 & \bf 50.4* & 31.5 & \bf 65.0*& 49.7 & \bf 51.1* & 32.1 \\
& \delibattobj & \bf 61.3* & 45.4 & \bf 47.9*$\dagger$ & 29.4 & \bf 67.7* & 52.6 & \bf 50.5* & 31.7 & \bf 65.0* & 49.5 & \bf 50.9* & 32.2\\
\bottomrule
\end{tabular}
}
\end{center}
\caption{\label{table:gapped_res} Results for the test sets 2016 and 2018 for the three degradation configurations: \rnd, \amb{} and \pers. M denotes \meteor, B -- \bleu; * marks statistically significant changes as computed for \meteor (p-value $\leq 0.05$) as compared to \base, $\dagger$ -- as compared to \delib. Bold highlights statistically significant improvements over \base.}
\end{table*}

\begin{figure*}[ht]
\small{
  \begin{subfigure}[c]{\textwidth}
  \vspace{1em}
    \begin{tabular}{c p{1.5cm}p{9cm}}
      \multirow{3}[15]{*}{\includegraphics[width=0.22\textwidth]{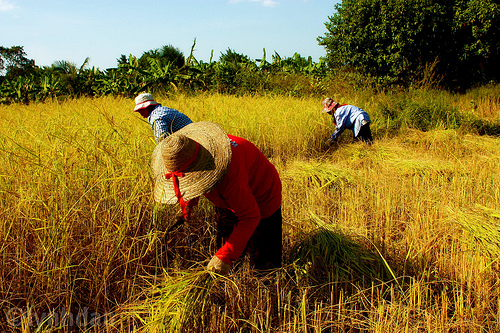}} & EN: & Three farmers harvest rice out in a rice \underline{field}.\\[1ex]
      & \base: & Drei Bauern ernten sich mit einem \underline{Reisfeld}.  \\[1ex]
      & \delib: & Drei Bauern ernten Reis mit einem \underline{Reisfeld}.\\[1ex]
      & \delibattobj: & Drei Bauern ernten sich mit einem \underline{Reishut} auf. \\[1ex]
      & DE: & Drei Farmer ernten Reis auf einem \underline{Feld}. \\[1ex]
  \end{tabular}
  \caption{Example of a blank resolved by the textual context for \amb: \textit{field} translated as \textit{Reisfeld} (rice field) by \base. \delibattobj incorrectly translated the blank into \textit{Reishut} (rice hat) due to detected objects. \textbf{Objects}: person, clothing, mammal}
  \end{subfigure}
  \begin{subfigure}[c]{\textwidth}
  \vspace{1em}
    \begin{tabular}{c p{1.5cm}p{9cm}}
      \multirow{3}[15]{*}{\includegraphics[width=0.22\textwidth]{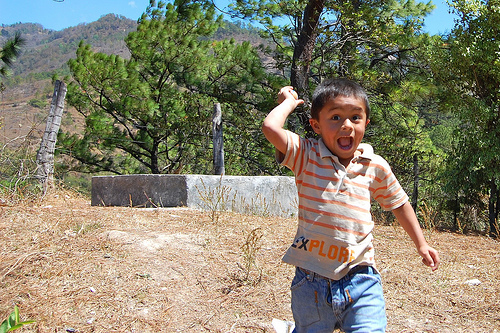}} & EN: & The \underline{boy} is outside enjoying a summer day.\\[1ex]
      & \base: & \underline{L'homme} profite d'une journ\'{e}e d'\'{e}t\'{e}.  \\[1ex]
      & \delib: & \underline{La femme} profite d'une journ\'{e}e d'\'{e}t\'{e}. \\[1ex]
      & \delibattobj: & \underline{L'enfant} profite d'une journ\'{e}e d'\'{e}t\'{e}. \\[1ex]
      & FR: & \underline{Le gar\c{c}on} est dehors, profitant d'une journ\'{e}e d'\'{e}t\'{e}. \\[1ex]
  \end{tabular}
  \caption{Example of a blank resolved by the multimodal context for \pers. The textual context is too generic and \delibattobj uses the detected objects to correctly translate \textit{boy} into \textit{l'enfant} (child). {\bf Objects}: clothing, face, tree, boy, jeans}
  \end{subfigure}
  \begin{subfigure}[c]{\textwidth}
  \vspace{1em}
    \begin{tabular}{c p{1.5cm}p{9cm}}
      \multirow{3}{*}{\includegraphics[width=0.22\textwidth]{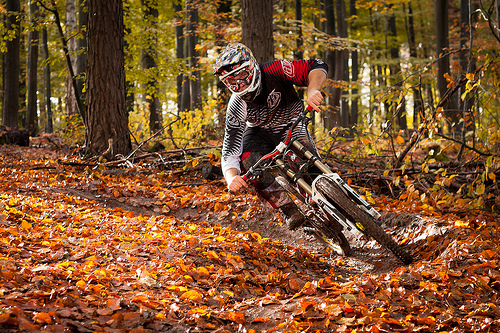}} & 
      EN: & Dirt \underline{biker} makes a sloping turn in a forest during the fall. \\[1ex]
      & \base: & \underline{Gel\"{a}ndemotorradfahrer} macht in einem Wald eine Kurve.\\[1ex]
      & \delib: & \underline{Gel\"{a}ndemotorradfahrer} macht in einem Herbst w\"{a}hrend Zuschauer eine Kurve.\\[1ex]
      & \delibattobj: & \underline{Gel\"{a}ndemotorradfahrer} macht in einem Herbst eine Kurve. \\[1ex]
      & DE: & Ein \underline{offroad-biker} f\"{a}hrt im Herbst durch eine steile Kurve.  \\[1ex]
  \end{tabular}
  \caption{Example of a blank resolved by the textual context for \pers. \textit{biker} correctly translated into the Masc. form \textit{Gel\"{a}ndemotorradfahrer} (dirt biker) by \base. \textbf{Objects}: person, tree, bike, helmet}
  \end{subfigure}}
 \caption{\label{table:gapped_ex} Examples of resolved blanks for test set 2016. Underlined text denotes blanked words and their translations. Object field indicates the detected objects.}
\end{figure*}

\subsection{Source degradation setup}\label{ssec:gaps_results}

Results of our source degradation experiments are shown in Table~\ref{table:gapped_res}. A first observation is that -- as with the standard setup -- the performance of our deliberation models is overall better than that of the base models. The results of the multimodal models differ for German and French. For German, \delibattobj is the most successful configuration and shows statistically significant improvements over \base for all setups. Moreover, for \rnd and \amb, it shows statistically significant improvements over \delib. However, especially for \rnd and \amb, \delib and \delibsum are either the same or slightly worse than \base. 

For French, all the deliberation models show statistically significant improvements over \base (average $\Delta\meteor=1$, $\Delta\bleu=1.1$), but the image information added to \delib only improve scores significantly for test 2018 \rnd. 

This difference in performances for French and German is potentially related to the need of more significant restructurings while translating from English into German.\footnote{English and French are both languages with the subject--verb--object (SVO) sentence structure. German, on the other hand, can have subject--object--verb (SOV) constructions. For example, a German sentence \textit{Gestern bin ich in London gewesen} ({\em Yesterday have I to London been}) would need to be restructured to \textit{Yesterday I have been to London} in English.} This is where a more complex \delibattobj architecture is more helpful. This is especially true for \rnd and \amb setups where blanked words could also be verbs, the part-of-speech most influenced by word order differences between English and German (see the decreasing complexity of translations for \delib and \delibattobj for the example (c) in Figure~\ref{table:gapped_ex}). 

To get an insight into the contribution of different contexts to the resolution of blanks, we performed manual analysis of examples coming from the English-German \base, \delib and \delibattobj setups (50 random examples per setup), where we count correctly translated blanks per system.

\begin{table}[!h]\begin{center}
\scalebox{0.73}{
\begin{tabular}{c c c c c}
\toprule
setup & \base & \delib & \delibattobj & gold \\ \midrule
\rnd & 22 & 23 & \bf 24 & 79 \\
\amb & 29 & 25 & \bf 33 & 88 \\
\pers & 43 & 46 & \bf 51 & 84\\
\bottomrule
\end{tabular}}
\end{center}
\caption{\label{table:gapped_human_res} Results of human annotation of blanked translations (English-German). We report counts of blanks resolved by each system, as well as total source blank count for each selection (50 sentences selected randomly).}
\end{table}

The results are shown in Table~\ref{table:gapped_human_res}. As expected, they show that the \rnd and \amb blanks are more difficult to resolve (at most 40\% resolved as compared to 61\% for \pers). Translations of the majority of those blanks tend to be guessed by the textual context alone (especially for verbs). Image information is more helpful for \pers: we observe an increase of 10\% in resolved blanks for \delibattobj as compared to \delib. However, for \pers the textual context is still enough in the majority of the cases: models tend to associate men with sports or women with cooking and are usually right (see Figure~\ref{table:gapped_ex} example (c)).

The cases where image helps seem to be those with rather generic contexts: see Figure~\ref{table:gapped_ex} (b) where \textit{enjoying a summer day} is not associated with any particular gender and make other models choose \textit{homme} (man) or \textit{femme} (woman), and only \baseattobj chooses \textit{enfant} (child) (the option closest to the reference).  

In some cases detected objects are inaccurate or not precise enough to be helpful (e.g., when an object Person is detected) and can even harm correct translations. 

\section{Conclusions} \label{sec:concl}
We have proposed a novel approach to multimodal machine translation which makes better use of context, both textual and visual. 
Our results show that further exploring textual context through deliberation networks already leads to better results than the previous state of the art. Adding visual information, and in particular structural representations of this information, proved beneficial when input text contains noise and the language pair requires substantial restructuring from source to target. Our findings suggest that the combination of a deliberation approach and information from additional modalities is a promising direction for machine translation that is robust to noisy input. %
Our code and pre-processing scripts are available at \url{https://github.com/ImperialNLP/MMT-Delib}.
\section*{Acknowledgments}
The authors thank the anonymous reviewers for their useful feedback. This work was supported by the MultiMT (H2020 ERC Starting Grant No. 678017) and MMVC (Newton
Fund Institutional Links Grant, ID 352343575)
projects. We also thank the annotators for their valuable help.

\bibliography{acl2019}
\bibliographystyle{acl_natbib}
\newpage
\appendix
\section{Appendices}
\label{sec:appendix}
\begin{figure*}[ht!]
\small{
  \begin{subfigure}[c]{\textwidth}
    \begin{tabular}{c p{1.5cm}p{9cm}}
      \multirow{3}[15]{*}{\includegraphics[width=0.22\textwidth]{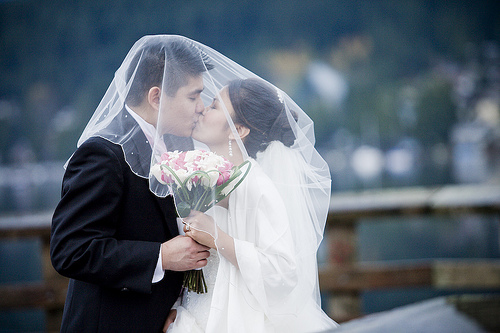}} & EN: & A \underline{bride} and \underline{groom} kiss under the \underline{bride}'s veil.\\[1ex]
      & \base: & Ein \underline{Mann} und eine \underline{Frau} k\"{u}ssen sich unter den Blicken der \underline{Frau}. \\[1ex]
      & \delib: & Ein \underline{Mann} und eine \underline{Frau} k\"{u}ssen sich unter dem \underline{Brautschleier}.\\[1ex]
      & \delibattobj: & Ein \underline{Mann} und eine \underline{Frau} k\"{u}ssen sich unter den \underline{hin}. \\[1ex]
      & DE: & Eine \underline{Braut} und \underline{Br\"{a}utigam} k\"{u}ssen sich unter dem \underline{Brautschleier} . \\[1ex]
  \end{tabular}
  \caption{\pers example: \textit{bride} and \textit{groom} translated are correctly translated by \base into \textit{Frau} (wife) and \textit{Mann} (husband). \textbf{Objects}: face, woman, dress}
  \end{subfigure}
  \begin{subfigure}[c]{\textwidth}
  \vspace{1em}
    \begin{tabular}{c p{1.5cm}p{9cm}}
      \multirow{3}[15]{*}{\includegraphics[width=0.10\textwidth]{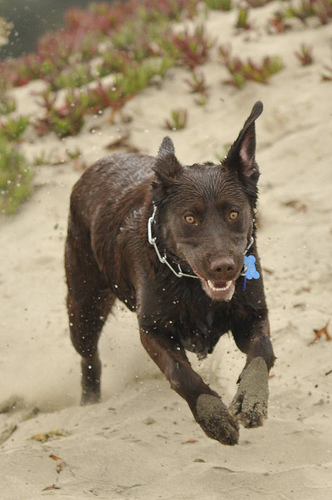}} & EN: & A brown dog \underline{runs} down the sandy beach.\\[1ex]
      & \base: & Ein brauner Hund \underline{l\"{a}uft} an einem sandigen Strand.\\[1ex]
      & \delib: & Ein brauner Hund \underline{rennt} den Sandstrand hinunter. \\[1ex]
      & \delibattobj: & Ein brauner Hund \underline{l\"{a}uft} an einem sandigen Strand hinunter. \\[1ex]
      & FR: & Ein brauner Hund \underline{l\"{a}uft} \"{u}ber den Sandstrand. \\[1ex]
  \end{tabular}
  \caption{\amb example: \textit{runs} is correctly translated by \base into \textit{l\"{a}uft}. {\bf Objects}: dog}
  \end{subfigure}
   }
 \caption{\label{table:gapped_ex1} Examples of blanks for test set 2016 that were correctly resolved by the textual context. The underlined words denote blanked words and their translations.}
\end{figure*}
\newpage
\begin{figure*}[ht!]
\small{
  \begin{subfigure}[c]{\textwidth}
  \vspace{1em}
    \begin{tabular}{c p{1.5cm}p{9cm}}
      \multirow{3}[15]{*}{\includegraphics[width=0.22\textwidth]{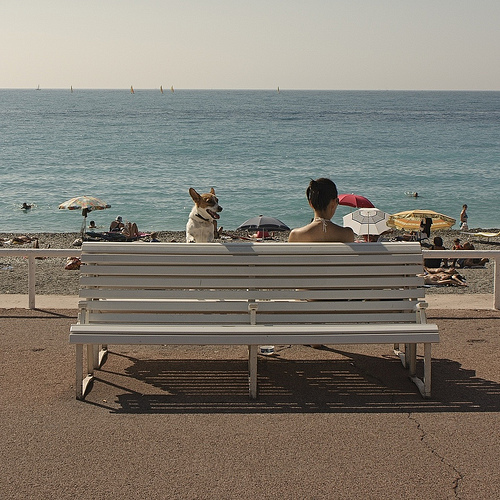}} & EN: & A woman and a dog sit on a white \underline{bench} near a beach.\\[1ex]
      & \base: & Eine Frau und ein Hund sitzen an einem wei{\ss}en \underline{Strand} nahe einem Strand. \\[1ex]
      & \delib: & Eine Frau und ein Hund sitzen auf einem wei{\ss}en \underline{Sofa} in der n\"{a}he eines Strands. \\[1ex]
      & \delibattobj: & Eine Frau und ein Hund sitzen auf einer wei{\ss}en \underline{Bank} nahe einem Strand. \\[1ex]
      & DE: & Eine Frau und eine Hund sitzen auf einer wei{\ss}en \underline{Bank} in der n\"{a}he eines Strandes.\\[1ex]
  \end{tabular}
  \caption{\rnd example: the blank \textit{bench} is correctly translated by \delibattobj into \textit{Bank} due to the detected object Bench. \textbf{Objects}: person, dog, bench}
  \end{subfigure}
  \begin{subfigure}[c]{\textwidth}
  \vspace{1em}
    \begin{tabular}{c p{1.5cm}p{9cm}}
      \multirow{3}{*}{\includegraphics[width=0.17\textwidth]{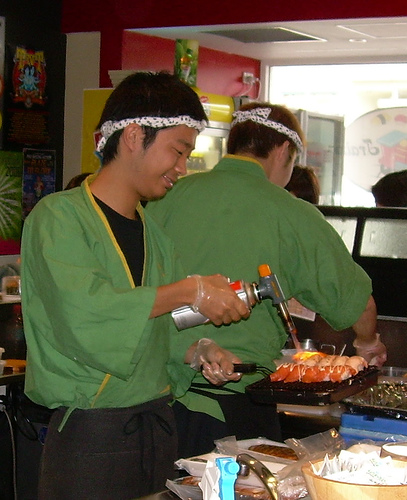}} & 
      EN: & Two \underline{men} dressed in green are preparing food in a restaurant.\\[1ex]
      & \base: & Deux \underline{femmes} v\^{e}tues de vert pr\'{e}parent des aliments dans un restaurant. \\[1ex]
      & \delib: & Deux \underline{femmes} v\^{e}tues de vert pr\'{e}parent de la nourriture dans un restaurant.\\[1ex]
      & \delibattobj: & Deux \underline{asiatiques} en vert pr\'{e}parent de la nourriture dans un restaurant. \\[1ex]
      & FR: & Deux \underline{hommes} habill\'{e}s en vert pr\'{e}parent de la nourriture dans un restaurant. \\[1ex]
  \end{tabular}
  \caption{\pers example. \textit{men} correctly translated into \textit{asiatiques} (asians) by \delibattobj. \textbf{Objects}: person, clothing, man, food, cake}
  \end{subfigure}}
 \caption{\label{table:gapped_ex2} Examples of blanks for test set 2016 that were correctly resolved by the multimodal context. The underlined words denote blanked words and their translations.}
\end{figure*}

\begin{figure*}[ht]
\small{
  \begin{subfigure}[c]{\textwidth}
  \vspace{1em}
    \begin{tabular}{c p{1.5cm}p{9cm}}
      \multirow{3}[15]{*}{\includegraphics[width=0.2\textwidth]{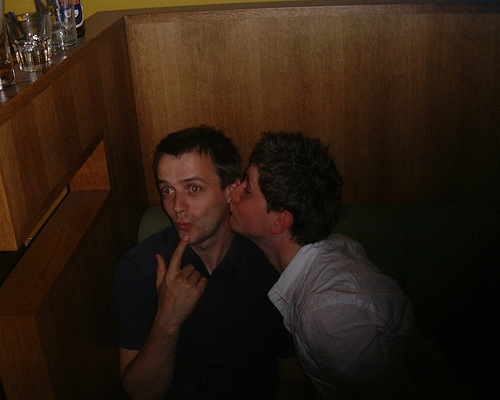}} & EN: & A \underline{guy} give a kiss to a \underline{guy} also.\\[1ex]
      & \base: & Ein \underline{Mann}, der sich vor, um eine \underline{Frau} zu kn\"{u}ssen .  \\[1ex]
      & \delib: & Ein \underline{Mann}, der sich vor, um eine \underline{Frau} zu k\"{u}ssen.  \\[1ex]
      & \delibattobj: & Ein \underline{Mann}, der einem kuss k\"{u}sst, um eine \underline{Frau} zu k\"{u}ssen. \\[1ex]
      & DE: & Ein \underline{Typ} k\"{u}sst einen anderen \underline{Typ} .\\[1ex]
  \end{tabular}
  \caption{\pers example: the second mention of \textit{guy} is consistently translated into \textit{Frau} (woman). \textbf{Objects}: clothing, man, face}
  \end{subfigure}
  \begin{subfigure}[c]{\textwidth}
  \vspace{1em}
    \begin{tabular}{c p{1.5cm}p{9cm}}
      \multirow{3}{*}{\includegraphics[width=0.17\textwidth]{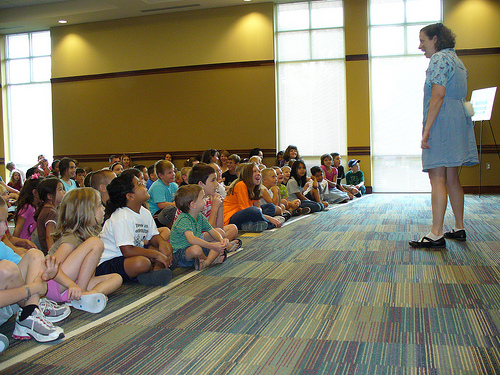}} & 
      EN: & A group of students sit and \underline{listen} to the \underline{speaker}.\\[1ex]
      & \base: & Eine Gruppe von Studenten sitzt und \underline{schaut} nach \underline{rechts} . \\[1ex]
      & \delib: & Eine Gruppe Sch\"{u}ler sitzt und \underline{schaut} nach \underline{rechts}. \\[1ex]
      & \delibattobj: & Eine Gruppe Sch\"{u}ler sitzt und \underline{schaut} zu rechts auf das \underline{Wasser}. \\[1ex]
      & DE: & Eine Gruppe von Studenten sitzt und \underline{h\"{o}rt} der \underline{Sprecherin} \underline{zu}.  \\[1ex]
  \end{tabular}
  \caption{\amb example. The blanks \textit{listen} and \textit{speaker} are consistently translated into \textit{schaut} (look) and \textit{rechts} (right) or \textit{Wasser} (water). \textbf{Objects}: person, clothing, man, food, cake}
  \end{subfigure}}
 \caption{\label{table:gapped_ex3} Examples of unresolved blanks. The underlined words denote blanked words and their translations.}
\end{figure*}
\end{document}